\documentclass[11pt,a4paper]{article}
\usepackage[margin=24mm]{geometry}
\usepackage[T1]{fontenc}
\usepackage[utf8]{inputenc}
\usepackage{lmodern}
\usepackage{amsmath,amssymb,booktabs,longtable,array,graphicx,xcolor,tikz,microtype}
\usetikzlibrary{arrows.meta,positioning,calc}
\usepackage[numbers,sort&compress]{natbib}
\usepackage[unicode,colorlinks=true,linkcolor=black,citecolor=black,urlcolor=blue!55!black]{hyperref}
\hypersetup{pdftitle={Vision Transformer Ensembles for Panoramic Street Segmentation},pdfauthor={Yunus Serhat Bıçakçı},pdfsubject={PalmCity challenge methods and measured results},pdfkeywords={semantic segmentation, panoramic street imagery, PalmCity, DINOv3, EoMT, ensembles, paired bootstrap, test time augmentation, input resolution}}
\definecolor{orcidgreen}{HTML}{A6CE39}
\definecolor{palmgreen}{HTML}{276B5B}
\definecolor{palmblue}{HTML}{345B88}
\newcommand{\orcidicon}{\href{https://orcid.org/0000-0002-7288-9959}{\raisebox{-0.12em}{\begin{tikzpicture}[x=1em,y=1em]\fill[orcidgreen] (0,0) circle (0.48);\node[white,font=\sffamily\bfseries\tiny,inner sep=0pt] at (0,0) {iD};\end{tikzpicture}}}}
\newcommand{\pct}{\%}
\newcommand{\code}[1]{\texttt{#1}}
\title{Vision Transformer Ensembles for Panoramic Street Segmentation}
\author{Yunus Serhat Bıçakçı\,\orcidicon\\[5pt]\small Department of Artificial Intelligence and Machine Learning\\\small Faculty of Applied Sciences, Marmara University\\\small Istanbul, Türkiye\\[5pt]\small Geospatial Data Science Group\\\small School of Geographical and Earth Sciences\\\small University of Glasgow, Glasgow, UK}
\date{5 October 2026, revised 30 October 2026}

\begin{document}
\maketitle
\begin{abstract}
Semantic segmentation of street panoramas can support detailed descriptions of urban environments, yet small datasets and unequal training costs make model selection difficult. This paper documents the two stages of a submission to the PalmCity challenge. In the first stage, nine pretrained segmentation systems are compared using approximately equal computation budgets, and the two leading candidates are trained with three random seeds. An equal probability average of three Encoder only Mask Transformer models with DINOv3 features, evaluated at three image scales with horizontal reflection, reaches 60.95\pct\ mean intersection over union on the 84 image validation split and 57.08\pct\ on the hidden test leaderboard, first in the snapshot of 5 October 2026. In the second stage, color and crop augmentation, rare class sampling, larger inputs, further architectures, and post processing are tested under the same update budget, and candidates are compared through a paired image bootstrap with selection rules fixed before the results. Training at 768 by 1536 and 896 by 1792 pixels gives the largest gains. The selected ensemble of four models scores 64.53\pct\ on validation when trained on the training images alone. Retrained on the training and validation images, it receives 61.54\pct\ mean intersection over union and 71.93\pct\ mean F1 on the hidden test leaderboard, first among five entries in the snapshot of 30 October 2026. Segment Anything voting, instance mask overlays, and ensembles with other architectures all reduce the validation score. Because the validation images guided many choices and the final models were trained on them, the validation scores are optimistic and the test score is the only independent measurement. The study reports all runs, comparisons, costs, and limits, and releases the code and models.
\end{abstract}

\section{Introduction}
Urban environments are experienced from the street. A map may show that a park, a road, or a building exists, while an image taken at pedestrian height reveals how vegetation, traffic, footpaths, street furniture, and physical barriers share the same view. These details matter when describing the quality and accessibility of public space. Street imagery has therefore become a useful complement to remote sensing and conventional spatial data. The review by Biljecki and Ito describes its applications across urban planning, transport, environmental assessment, and related fields \citep{biljecki2021review}. The growing variety of these applications creates a practical need for segmentation systems whose output can be inspected, reproduced, and interpreted at the class level.

The connection between visible surroundings and urban outcomes requires care. Helbich and colleagues examined associations between green and blue space measured from street imagery and depression symptoms among older adults in Beijing \citep{helbich2019greenblue}. Their later Amsterdam study compared overhead and eye level greenness measures and found no statistically significant association with depression or anxiety scores \citep{helbich2021greenness}. The studies show that both the measurement perspective and the setting matter. A street facing camera and an overhead sensor do not necessarily represent the same exposure. At the same time, a segmentation mask is only one component of an environmental measurement process. Sampling locations, image dates, geographical coverage, model errors, and the design of any subsequent health study also affect its conclusions.

Street imagery can support more specific analyses when its visual categories are linked to a clearly defined question. Ito and Biljecki used street imagery and computer vision to assess bikeability \citep{ito2021bikeability}. Recent work in the Fatih district of Istanbul used vision language descriptions and multimodal embeddings from 116696 Mapillary images to identify interpretable urban typologies without manual labels \citep{bicakci2026fatih}. This provides another example of extracting spatially meaningful information from street observations. These applications motivate reliable visual measurements. PalmCity supplies semantic labels for image regions, and the present evaluation concerns the accuracy and cost of recovering those labels.

Much of the public evidence used to develop computer vision systems comes from a limited set of places and imaging conditions. A model that works well on one benchmark may encounter different streets, vehicles, surfaces, vegetation, or camera geometry elsewhere. PalmCity addresses part of this gap through panoramic street imagery captured in Mersin, Türkiye, and a benchmark motivated by underrepresented developing countries \citep{iban2026palmcity}. Its 32 categories cover both large regions such as road, building, sky, and vegetation, and small objects such as signs, traffic lights, benches, and infrastructure components. Panoramic imagery also places wide context and fine details within the same image. A useful challenge system must handle both.

The size of the labelled training split creates a second difficulty. There are only 497 training panoramas. Modern pretrained networks offer powerful visual features, but their performance after adaptation depends on the original training data, decoder, optimization settings, and the amount of computation available. Simply training every candidate for the same number of epochs gives slower systems a larger computation allowance. Conversely, a very short common schedule may fail to reveal the value of a large model. An evaluation that records both quality and cost is more informative than one that reports only the largest validation score.

Recent segmentation systems provide several plausible routes. DeepLabV3+ combines convolutional features with a decoder for spatial detail \citep{chen2018deeplab}. SegFormer uses a hierarchical transformer and a lightweight segmentation head \citep{xie2021segformer}. UPerNet combines features at several resolutions \citep{xiao2018upernet}, and can be paired with Swin Transformer \citep{liu2021swin} or ConvNeXt \citep{liu2022convnext}. Mask2Former frames segmentation through learned mask queries \citep{cheng2022mask2former}. DINOv3 supplies visual representations that can be adapted to dense prediction \citep{simeoni2025dinov3}. The Encoder only Mask Transformer, abbreviated EoMT, places mask queries within a vision transformer and provides an additional candidate \citep{kerssies2025eomt}. Choosing among these alternatives is an empirical question under the resources available for the challenge.

Prediction fusion offers another practical choice after selecting a base model. Lakshminarayanan and colleagues describe deep ensembles that average predicted probabilities from independently trained neural networks \citep{lakshminarayanan2017deepensembles}. Fusion takes different forms across vision tasks. In a 2026 preprint, Bıçakçı combines CNN and transformer detections in aerial imagery through Weighted Boxes Fusion \citep{bicakci2026hybridensemble}. Ozturk and colleagues average the softmax outputs of four CNN architectures for tea leaf disease classification \citep{ozturk2025teaensemble}. In the selected PalmCity ensemble, members share one architecture and differ in their training seed. Their aligned class probabilities are averaged at each panorama pixel. The benefit is assessed through PalmCity experiments against individual models and alternative inference settings.

This paper follows the complete path from candidate selection to the submitted mask archive in two stages. In the first stage, a pilot compares nine systems using an estimated 600 seconds of training and validation computation per system. The two leading systems receive an estimated 2400 seconds each for three independent random seeds. A fixed set of eleven inference variants then compares individual checkpoints, probability ensembles, image scales, horizontal reflection, and overlapping windows. The final choice uses the official mean intersection over union, with class level performance, runtime, and memory providing additional evidence.

The second stage began after the first leaderboard snapshot. On 20 October 2026 another participant submitted an entry with 60.55\pct\ mean intersection over union, above the first submission \citep{codabench2026second}. Between 28 and 30 October the selected EoMT recipe was revisited under the same budget of 6351 optimizer updates. The second stage tests color and crop augmentation, rare class image sampling, more points for query matching, larger input sizes, a longer Mask2Former schedule, SegFormer and YOLO26 models initialized from Cityscapes \citep{cordts2016cityscapes,jocher2026yolo26}, EoMT initialized from Cityscapes, two post processing methods, and larger ensembles. Candidates are compared with a paired image bootstrap \citep{efron1979bootstrap}, and the selection rules are written into the experiment scripts before the corresponding results exist. The chosen recipes are finally retrained on the training and validation images together.

The first submission is based on three EoMT models with DINOv3 ViT L features. On the public validation split it reaches 60.95\pct\ mean intersection over union. The hidden test score shown on Codabench is 57.08\pct, which placed the submission first in the leaderboard snapshot on 5 October 2026. The second stage submission averages four EoMT models trained at 768 by 1536 and 896 by 1792 pixels and receives 61.54\pct, first among five entries in the snapshot of 30 October 2026. The measured selection procedure is the central contribution. The published artifacts retain unsuccessful alternatives and classes on which the selected systems remain weak. This emphasis on transparent and reusable procedures is consistent with recent recommendations for repeatable, reproducible, and expandable geospatial analysis \citep{biljecki2026reproducible}.

\section{Benchmark and evaluation}
\subsection{Data and split checks}
The experiments use the official PalmCity split without modification. Table~\ref{tab:data} gives its size. Every panorama has a width of 1024 pixels and a height of 512 pixels. Training and validation annotations are single channel images with direct integer class identifiers from 0 to 31. The labels are preserved throughout loading, training, evaluation, and archive construction. No label reduction inherited from an ADE20K configuration is applied. All comparisons in both stages use this split. Only the final models of the second stage are trained on a manifest that appends the 84 validation images to the 497 training images, as described in Section~\ref{sec:trainval}.

\begin{table}[htbp]
\centering
\caption{Official split used in the experiments. Hidden test annotations were not available to the training or selection procedure.}
\label{tab:data}
\begin{tabular}{lrr}
\toprule
Split & RGB panoramas & Available masks\\
\midrule
Training & 497 & 497\\
Validation & 84 & 84\\
Hidden test & 249 & 0\\
\bottomrule
\end{tabular}
\end{table}

The dataset audit checks file counts, image modes, dimensions, class ranges, and exact duplicates across splits. All 32 categories occur in both training and validation. No identical decoded RGB panorama occurs in more than one split. Some visually similar views nevertheless occur across the official training and validation splits. One inspected pair shows the same coastal walkway and nearby buildings from different viewpoints. This distinction between file separation and geographical separation is considered in the discussion.

Class support is highly unequal. The eight least frequent classes, selected from the training pixel histogram before the inference comparison, are Pruned Tree, Traffic Light, Driver, Bicycle, Traffic Sign, Sitting Bench, Railing, and Overpass. Their average IoU is reported as an additional diagnostic. It is not a substitute for the official metric. In validation, Overpass occupies only 40 labelled pixels in one image and Pruned Tree occupies 54 pixels in one image. Traffic Light is also present in only one validation image. These categories can have unstable scores even when the larger scene regions are predicted well.

\subsection{Exact scoring convention}
Evaluation accumulates a single confusion matrix over all pixels in the relevant split. Let $C_{ij}$ be the number of pixels whose true class is $i$ and predicted class is $j$. For class $c$, define $T_c=C_{cc}$, $P_c=\sum_i C_{ic}$, and $G_c=\sum_j C_{cj}$. The class scores are
\begin{equation}
\operatorname{IoU}_c=\frac{T_c}{P_c+G_c-T_c},\qquad
\operatorname{F1}_c=\frac{2T_c}{P_c+G_c}.
\label{eq:metric}
\end{equation}
A score with a zero denominator is set to zero. Mean IoU and mean F1 are the arithmetic means over all 32 classes. Class 31, named Void, remains part of both training and evaluation. A class absent from both the truth and the predictions contributes zero rather than being omitted. The metric is not an average of image specific mean IoU values.

This detail matters when comparing results with implementations that exclude Void or ignore absent categories. The local scorer was checked against the published PalmCity scoring program on 103 synthetic confusion matrices, with exact agreement. The eleven validation output sets were also independently decoded and scored using dataset level confusion matrices. This establishes agreement with the published scorer. It does not claim that the local code is byte identical to the private Codabench evaluation package.

\subsection{Paired image bootstrap}
\label{sec:bootstrap}
Differences between candidates on 84 images can be small, so the second stage reports a paired image bootstrap \citep{efron1979bootstrap} with each comparison. One confusion matrix is stored for every validation image. Each of 5000 replicates draws 84 images with replacement, sums the matrices of the drawn images separately for the candidate and the baseline, and recomputes the official mean IoU of both. The reported interval spans the 2.5 and 97.5 percentiles of the replicate differences. The positive share is the fraction of replicates in which the candidate scores higher. The random generator uses the fixed seed 20261028. The interval describes the sampling of these validation images only. It ignores the correlation between nearby views and the repeated use of the same images for selection, so it describes uncertainty without serving as a significance test.

\section{Candidate systems and transfer sources}
The reference systems are DeepLabV3+ with ResNet50 and SegFormer with MiT B2. The larger candidates are SegFormer B5, UPerNet with ConvNeXt Large or Swin Large, Mask2Former with Swin Large, DINOv3 ViT B or ViT L with a linear segmentation head, and EoMT with DINOv3 ViT L. Table~\ref{tab:pilots} lists the eligible results.

The initialization sources differ. The ResNet50 and MiT B2 reference encoders use ImageNet weights with newly initialized segmentation heads. SegFormer B5, UPerNet, Mask2Former, and EoMT start from complete semantic segmentation checkpoints trained on ADE20K \citep{zhou2019ade20k}. The DINOv3 linear systems use pretrained DINOv3 encoders and a newly initialized segmentation head. Every PalmCity class prediction layer is adapted to 32 categories. All encoders are updated during PalmCity training.

The linear DINOv3 decoder provides a simple way to test the usefulness of the representation. The final normalized patch features are reshaped to the actual image grid and passed through dropout, batch normalization, and a $1\times1$ convolution. Predictions are interpolated to the input size. Class and register tokens are excluded from the patch grid.

EoMT is an especially relevant addition because it couples strong pretrained features with mask prediction in the encoder. The implementation uses the available DINOv3 ViT L ADE20K semantic checkpoint from the official model release \citep{eomt2026dinov3zoo}. This later release is distinguished from the original EoMT paper, which evaluates DINOv2 based systems. Its mask reshape grid is derived from the actual rectangular image dimensions on each forward pass to preserve the panorama layout. The training schedule records the decay of query attention masks. These masks are disabled during validation and inference, following the intended evaluation behavior.

The selected EoMT configuration has 24 transformer layers, a feature width of 1024, 16 attention heads, and patches of $16\times16$ pixels. A native panorama therefore supplies a $32\times64$ patch grid. One hundred learned mask queries are used in the final four transformer blocks. The query classification, mask, and Dice loss weights are 2, 5, and 5, with a no object weight of 0.1. Query attention mask probabilities decrease polynomially over four recorded update intervals, spanning fractions 0 to 0.2, 0.4 to 0.6, 0.6 to 0.8, and 0.8 to 1 of the training budget. The exponent is 0.9. These settings are included in the released configuration.

Native losses are retained when they are integral to a system. UPerNet uses cross entropy with its auxiliary supervision. Mask2Former and EoMT use their native mask and class objectives with query matching. DeepLabV3+, SegFormer, and the linear DINOv3 systems use cross entropy with a Dice term. For query systems, semantic targets contain one binary mask per class present in an image, including Void. Their additional no object query category is distinct from PalmCity class 31.

Pretrained sources and checkpoint versions are recorded in the public repository. Access conditions and checkpoint licences are documented separately from the code licence. The challenge workflow permits pretrained weights and external data. PalmCity remains the only dataset directly used for the adaptation runs, while the transferred weights retain information learned from their upstream datasets. Section~\ref{sec:second} describes the further systems of the second stage.

\section{A bounded training protocol}
\subsection{Calibration and computation budgets}
Every candidate is calibrated using real PalmCity data at the full $512\times1024$ input size. Calibration uses one warmup optimizer update and three measured updates, together with one warmup inference and three measured validation images. Update timings include decoding, augmentation, loading, device transfer, forward and backward computation, gradient clipping, and AdamW. Model initialization and checkpoint writing are recorded separately. A short calibration estimates steady state cost and is not treated as evidence of segmentation quality.

All models use BF16 mixed precision and an effective batch size of four. Microbatch sizes and gradient accumulation vary with the model. DeepLabV3+ and UPerNet use microbatches of two because their pooled batch normalization requires more than one sample. SegFormer B2 also uses microbatches of two. The remaining systems use microbatches of one with four accumulation steps. The linear DINOv3 systems use gradient checkpointing. Accumulation correctly weights the number of samples in the final partial group.

The pilot budget is an estimated 600 seconds of training plus validation computation for each candidate. Calibration translates this estimate into a fixed optimizer update limit, maximum epoch count, and warmup length. Ten validation evaluations are reserved in the budget and determine the checkpoint with the highest mean IoU. Fixed update limits and variation in measured update and validation speed cause the computation time to differ from the estimate. Eligible pilots use 603.15 to 628.02 seconds of measured training and validation computation. Initialization and checkpoint writing add to the whole process duration and are excluded from these figures. The comparison therefore uses approximately equal computation.

The top two pilot systems are then trained from their original pretrained sources, rather than by continuing the pilot weights. Each receives an estimated 2400 seconds for seeds 42, 123, and 2026. EoMT has a limit of 6351 updates and 51 epochs, while DINOv3 ViT L with a linear head has 7940 updates and 64 epochs. Each run again has ten validation evaluations. All six runs finish within 1.81\pct\ to 4.55\pct\ above the computation target. EoMT and DINOv3 confirmations run on separate 32 GB RTX 5090 cards. No model combines the memory of the two cards.

\begin{figure}[htbp]
\centering
\begin{tikzpicture}[node distance=7mm and 11mm, every node/.style={font=\small}, stage/.style={draw=palmgreen,rounded corners=2pt,align=center,text width=6.7cm,minimum height=15mm,inner sep=5pt}, flow/.style={-{Stealth[length=2mm]},thick,palmgreen}]
\node[stage] (a) {Official data and 32 class scoring\\497 train, 84 validation, 249 test};
\node[stage,right=of a] (b) {Nine eligible transfer systems\\600 seconds estimated per pilot};
\node[stage,below=of b] (c) {Two candidates and three seeds each\\2400 seconds estimated per run};
\node[stage,left=of c] (d) {Eleven validation inference variants\\Models, ensembles and transforms};
\node[stage,below=of d,xshift=3.9cm,text width=14.6cm] (e) {Selected three seed EoMT ensemble with scales and reflection\\Strict archive containing 249 native class ID masks};
\draw[flow] (a)--(b);\draw[flow] (b)--(c);\draw[flow] (c)--(d);\draw[flow] (d.south)--(d.south |- e.north);
\end{tikzpicture}
\caption{The selection procedure uses public validation labels throughout. Hidden test labels are never used for training, checkpoint selection, or inference variant selection.}
\label{fig:workflow}
\end{figure}
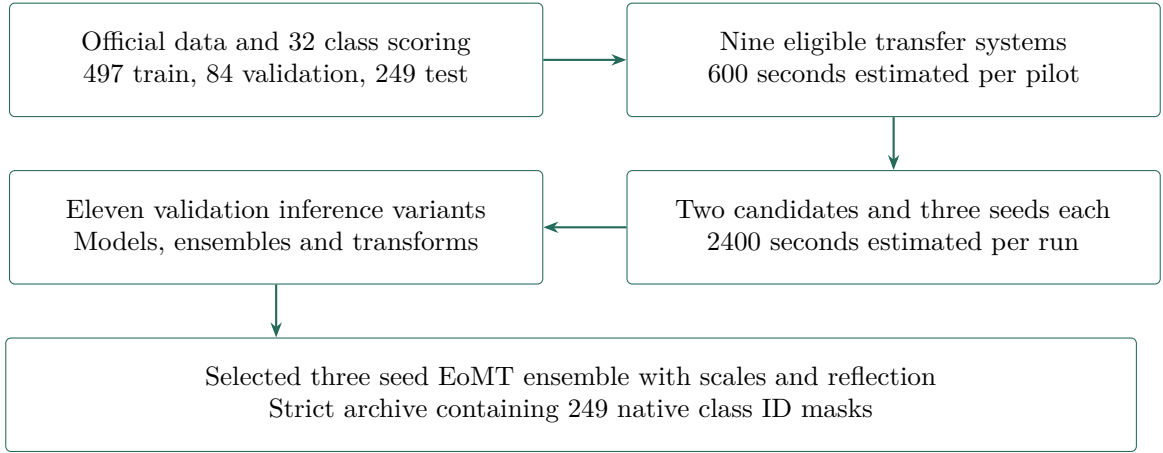

\subsection{Optimization and recorded recipe corrections}
All runs use AdamW \citep{loshchilov2019adamw}. Bias and normalization parameters are excluded from weight decay. Learning rates use a polynomial schedule with power 0.9, a minimum value of $10^{-6}$, and warmup occupying 5\pct\ of the update budget. The data augmentation is a horizontal reflection with probability 0.5. There are no random crops, panorama rolls, or changes to the $2\mathbin{:}1$ aspect ratio. Training remains at the native spatial resolution. The second stage changes the augmentation and the input size, as described in Section~\ref{sec:second}.

Table~\ref{tab:optimizer} gives the eligible recipes. These are adaptations for a short PalmCity fine tuning run, rather than reproductions of the full source training schedules. EoMT uses a common learning rate of $6\times10^{-5}$ for the encoder and prediction layers. Its original ADE20K schedule uses a larger base learning rate and a longer training procedure. ConvNeXt uses one scalar encoder multiplier as an approximation to the source stage dependent values. These choices limit implementation and search complexity, and their effect remains part of the evaluated configuration.

\begin{table}[htbp]
\centering
\small
\caption{Eligible optimizer settings. The encoder learning rate equals the base rate multiplied by the encoder factor. Gradient clipping uses the recorded norm threshold.}
\label{tab:optimizer}
\begin{tabular}{lrrrr}
\toprule
System & Base learning rate & Encoder factor & Decay & Clip\\
\midrule
DeepLabV3+ ResNet50 & $10^{-4}$ & 1 & 0.01 & 1\\
SegFormer B2 and B5 & $6\times10^{-4}$ & 0.1 & 0.01 & 1\\
DINOv3 linear B and L & $6\times10^{-4}$ & 0.1 & 0.01 & 1\\
EoMT DINOv3 L & $6\times10^{-5}$ & 1 & 0.01 & 1\\
UPerNet ConvNeXt L & $10^{-4}$ & 0.3 & 0.05 & 1\\
UPerNet Swin L & $6\times10^{-5}$ & 1 & 0.01 & 1\\
Mask2Former Swin L & $10^{-4}$ & 0.1 & 0.05 & 0.01\\
\bottomrule
\end{tabular}
\end{table}

An initial set of nine pilots used a more uniform adapted recipe. Review of the source settings revealed four changes worth applying before the primary comparison. The changes concern the SegFormer encoder and head learning rate ratio, the Mask2Former encoder rate, decay and clipping, and the ConvNeXt encoder rate and decay. These four configurations were rerun under the same pilot budget. The primary cohort uses the specified corrected recipes and the five unchanged original runs. It does not select the larger score from each old and corrected pair. In particular, the corrected Mask2Former score is slightly lower than its original score and remains the eligible result. Both versions are retained in the experiment record. There are therefore 13 pilot executions and six confirmation executions, giving 19 training runs in total.

The complete recorded training and validation computation is 6.3443 GPU hours, including the four additional diagnostic pilot runs and excluding initialization and checkpoint input and output. The inference sweep takes another 392.63 seconds of elapsed time and finishes within its fixed 1800 second allowance.

\subsection{Reproducibility controls}
Before training, source code and dependencies are frozen. Each run records the resolved model configuration, pretrained source, dataset inputs, random seed, update limits, validation schedule, and measured resources. Checkpoints include model weights, optimizer and scheduler state, mixed precision state, and random number generator state. Resume operates at an epoch boundary and checks that the saved experiment settings match before proceeding.

The environment uses Python 3.12.12, PyTorch 2.9.1 with CUDA 12.8, torchvision 0.24.1, Transformers 5.5.2, segmentation models pytorch 0.5.0, timm 1.0.30, and SciPy 1.17.1. The dependency lock is retained, and the second stage uses the same lock. Random seeds are recorded, and strict CUDA determinism is disabled. Hardware and numerical differences can therefore affect a rerun.

\section{Probability ensembles and inference variants}
\subsection{Comparable semantic probabilities}
The inference pipeline converts every model output into a distribution over the same 32 semantic classes. For a dense predictor this is the softmax of its class logits. Query predictors require a different conversion. If $a_{qc}$ is the softmax probability that query $q$ has semantic class $c$, and $m_q(u)$ is the sigmoid mask value at pixel $u$, the unnormalized semantic score is
\begin{equation}
z_c(u)=\sum_q a_{qc}m_q(u),\qquad
p_c(u)=\frac{z_c(u)}{\sum_{k=0}^{31}z_k(u)}.
\label{eq:query}
\end{equation}
Only the separate no object query category is discarded. Void is retained. Small positive clamping prevents numerical failure before normalization. Returning the logarithm of the normalized probabilities allows the shared dense inference interface to recover the intended distribution through a softmax. Without this step, applying a softmax directly to positive mask scores would change the ensemble meaning.

For the selected system, let $p_{s,t,c}(u)$ be the probability from training seed $s$ and inference transform $t$, aligned back to the native image coordinates. The ensemble is
\begin{equation}
\bar p_c(u)=\frac{1}{3\cdot6}\sum_{s\in\{42,123,2026\}}\sum_{t\in\mathcal T}p_{s,t,c}(u),\qquad
\hat y(u)=\operatorname*{arg\,max}_{c\in\{0,\ldots,31\}}\bar p_c(u).
\label{eq:ensemble}
\end{equation}
The transform set $\mathcal T$ contains scales 0.75, 1, and 1.25, each with the original and horizontally reflected image. Reflection is reversed before averaging. Scale inputs preserve the panorama aspect ratio and correspond to heights 384, 512, and 640 pixels. Outputs are aligned to the original $512\times1024$ grid using bilinear interpolation. The pipeline averages probabilities, never integer class masks.

The second stage ensemble applies Equation~\eqref{eq:ensemble} to four models with different input sizes. With $\mathcal M$ the set of selected models,
\begin{equation}
\bar p_c(u)=\frac{1}{4\cdot6}\sum_{m\in\mathcal M}\sum_{t\in\mathcal T_m}p_{m,t,c}(u),
\label{eq:ensemble2}
\end{equation}
where $\mathcal T_m$ contains scales 1, 1.25, and 1.5 of the input size of model $m$, each with the original and the reflected image. The three models trained at $768\times1536$ pixels therefore see heights of 768, 960, and 1152 pixels, and the model trained at $896\times1792$ pixels sees heights of 896, 1120, and 1344 pixels. The native panorama is resampled to each size, and every output is aligned back to the native grid as before. The bound on the number of pixels per probability map is raised to four million so that the largest view fits.

There is a small implementation detail in the spatial alignment. Query mask logits are interpolated before sigmoid aggregation. The common inference interface then interpolates log probabilities when mapping the scaled result back to the native grid and normalizes them with softmax. This is the recorded implementation used for the reported scores. It should be retained when reproducing the output rather than silently replacing it with a different interpolation order.

\subsection{Fixed comparisons and memory control}
The inference plan evaluates eleven variants on the same 84 validation images. It includes the best EoMT checkpoint, the best linear DINOv3 checkpoint, three weights for their combination, an equal ensemble of the three EoMT seeds, reflection alone, three scales with reflection, and windows of $256\times512$ pixels with 0.5 overlap. The candidate list is fixed before the sweep. It is not expanded in response to individual scores.

Only one panorama is processed at a time. Probability sums remain on the CPU, and the ensemble models are moved to the GPU sequentially for each image. This permits three checkpoints to be used without keeping all model weights in GPU memory together. It also adds model transfer cost. Reported prediction time includes image decoding, tensor preparation, transfers, forward inference, transformation reversal, class selection, and PNG writing. Checkpoint loading and source checks are included in the separate total runtime. Allocated memory is measured through PyTorch and is not the complete memory usage reported by the graphics driver.

\section{Results}
\subsection{Eligible pilot comparison}
Table~\ref{tab:pilots} reports the nine eligible pilot results. EoMT has the highest mean IoU at 57.565\pct. The linear DINOv3 ViT L system reaches 52.151\pct\ and Mask2Former reaches 51.926\pct. The difference between the second and third candidate is only 0.226 percentage points. Selecting the second candidate for confirmation is therefore a bounded search decision, not strong evidence that Mask2Former cannot perform better with a different schedule.

\begin{table}[htbp]
\centering
\small
\caption{Eligible pilot results at approximately 600 seconds of training and validation computation. Scores are percentages. Rare IoU is the mean over the eight categories fixed from training pixel frequency. Memory is peak allocated training memory in GiB.}
\label{tab:pilots}
\begin{tabular}{lrrrrr}
\toprule
System & mIoU & mF1 & Rare IoU & Seconds & GiB\\
\midrule
EoMT DINOv3 L & 57.565 & 68.515 & 26.843 & 624.20 & 10.58\\
DINOv3 L linear & 52.151 & 63.566 & 24.841 & 628.02 & 5.71\\
Mask2Former Swin L & 51.926 & 62.885 & 15.591 & 609.14 & 9.36\\
DINOv3 B linear & 49.110 & 60.300 & 21.979 & 603.15 & 1.99\\
UPerNet ConvNeXt L & 48.784 & 59.043 & 13.314 & 609.52 & 9.59\\
UPerNet Swin L & 46.876 & 56.323 & 5.802 & 605.99 & 13.05\\
SegFormer B2 & 46.221 & 56.780 & 12.502 & 608.80 & 4.09\\
SegFormer B5 & 45.742 & 55.962 & 7.524 & 618.96 & 6.02\\
DeepLabV3+ ResNet50 & 42.306 & 51.539 & 5.057 & 619.87 & 2.43\\
\bottomrule
\end{tabular}
\end{table}

The pilot ranking does not simply follow model size. SegFormer B5 trails B2 under the eligible recipes and budget. DINOv3 ViT B with a linear decoder exceeds both UPerNet variants, while using less training memory. EoMT combines the highest score with the highest rare category mean in the pilot. These observations support the decision to continue EoMT and the linear DINOv3 ViT L system. They also show why a broad initial comparison is useful before investing in longer runs.

The recipe corrections have a visible effect. SegFormer B5 changes from 34.637\pct\ to 45.742\pct\ mean IoU, B2 changes from 44.323\pct\ to 46.221\pct, and ConvNeXt UPerNet changes from 47.747\pct\ to 48.784\pct. Mask2Former changes from 52.015\pct\ to 51.926\pct. The magnitude of the B5 change illustrates that a weak short run may reflect an unsuitable adaptation recipe. It does not justify selecting every source setting after seeing the result, which is why the correction cohort is explicitly retained.

\subsection{Longer training with three seeds}
The confirmation runs in Table~\ref{tab:seeds} strengthen the selection evidence. EoMT achieves a mean IoU of 58.791\pct\ across the three seeds, compared with 54.488\pct\ for the linear DINOv3 ViT L system. Every EoMT run exceeds every linear DINOv3 run. The mean difference is 4.303 percentage points. EoMT uses about 10.58 GiB of allocated training memory, compared with 5.71 GiB for the linear system.

\begin{table}[htbp]
\centering
\small
\caption{Independent confirmation runs with approximately 2400 seconds of training and validation computation each. The checkpoint is selected from ten scheduled validation evaluations per run.}
\label{tab:seeds}
\begin{tabular}{lrrrrr}
\toprule
System & Seed & mIoU & mF1 & Rare IoU & Best epoch\\
\midrule
EoMT DINOv3 L & 42 & 57.970 & 68.576 & 31.336 & 26\\
EoMT DINOv3 L & 123 & 58.731 & 69.353 & 30.846 & 36\\
EoMT DINOv3 L & 2026 & 59.670 & 69.718 & 31.103 & 31\\
DINOv3 L linear & 42 & 54.660 & 66.021 & 28.406 & 45\\
DINOv3 L linear & 123 & 54.706 & 65.862 & 29.475 & 52\\
DINOv3 L linear & 2026 & 54.098 & 65.523 & 27.448 & 58\\
\midrule
EoMT mean & & 58.791 & 69.216 & 31.095 &\\
DINOv3 linear mean & & 54.488 & 65.802 & 28.443 &\\
\bottomrule
\end{tabular}
\end{table}

The population standard deviation of mean IoU across seeds is 0.695 percentage points for EoMT and 0.276 for the linear DINOv3 system. These values describe observed training variation. The best individual EoMT checkpoint is seed 2026 at epoch 31. Its score of 59.670\pct\ is used as the individual reference in the inference comparison. The best linear DINOv3 checkpoint is seed 123 at epoch 52, with 54.706\pct.

\subsection{Inference selection and its cost}
The full inference comparison appears in Table~\ref{tab:inference}. Averaging the three EoMT seeds at the native image scale gives 59.852\pct\ mean IoU, a modest improvement of 0.181 percentage points over the strongest single checkpoint. Using three scales with reflection on the single checkpoint gives a larger improvement, reaching 60.671\pct. Combining the three seeds with the same transforms gives the highest validation score, 60.953\pct\ mean IoU and 71.165\pct\ mean F1.

\begin{table}[htbp]
\centering
\small
\caption{All eleven inference variants on 84 validation panoramas. E denotes the best EoMT checkpoint and D the best linear DINOv3 checkpoint. Ensemble weights apply to normalized semantic probabilities. Seconds measure the complete prediction loop, including transfers and PNG output.}
\label{tab:inference}
\begin{tabular}{lrrrr}
\toprule
Inference variant & mIoU & mF1 & Seconds & GiB\\
\midrule
Three EoMT seeds, scales and reflection & \textbf{60.953} & \textbf{71.165} & 81.72 & 2.70\\
E, scales and reflection & 60.671 & 70.828 & 16.77 & 2.70\\
Three EoMT seeds, native & 59.852 & 70.093 & 51.82 & 2.32\\
E, native & 59.670 & 69.718 & 7.85 & 2.32\\
$0.75E+0.25D$, native & 59.652 & 69.624 & 34.72 & 2.32\\
Three EoMT seeds, reflection & 59.510 & 69.595 & 56.95 & 2.39\\
E, overlapping windows & 59.070 & 69.809 & 13.68 & 1.91\\
E, reflection & 59.025 & 68.930 & 8.34 & 2.39\\
$0.5E+0.5D$, native & 58.364 & 68.725 & 34.78 & 2.32\\
$0.25E+0.75D$, native & 55.863 & 66.842 & 34.69 & 2.32\\
D, native & 54.706 & 65.862 & 6.96 & 1.79\\
\bottomrule
\end{tabular}
\end{table}

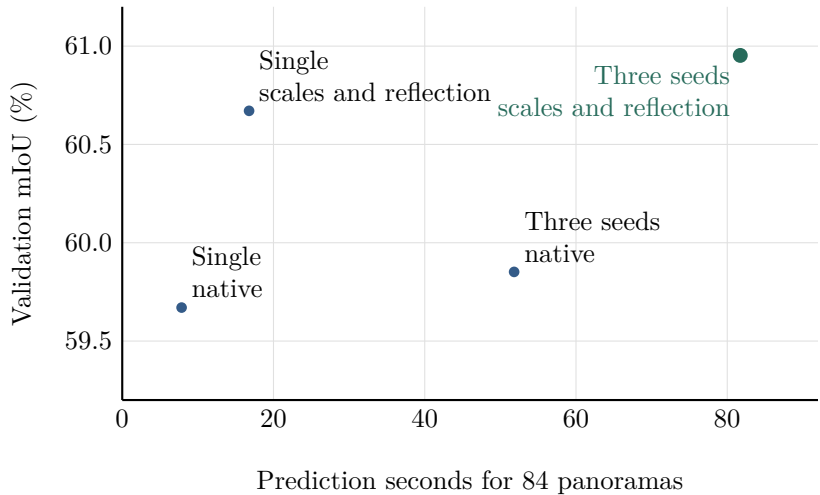
\begin{figure}[htbp]
\centering
\begin{tikzpicture}[x=0.10cm,y=2.6cm,every node/.style={font=\small}]
\foreach \x in {0,20,40,60,80} {\draw[black!12] (\x,0)--(\x,2);\node[below] at (\x,0) {\x};}
\foreach \y/\label in {0.3/59.5,0.8/60.0,1.3/60.5,1.8/61.0} {\draw[black!12] (0,\y)--(92,\y);\node[left] at (0,\y) {\label};}
\draw[thick] (0,0)--(92,0);\draw[thick] (0,0)--(0,2);
\node[below=8mm] at (46,0) {Prediction seconds for 84 panoramas};
\node[rotate=90] at (-13,1) {Validation mIoU (\%)};
\fill[palmblue] (7.84782,0.470276) circle[radius=2pt];
\node[above right,align=left] at (7.84782,0.470276) {Single\\native};
\fill[palmblue] (51.81796,0.651724) circle[radius=2pt];
\node[above right,align=left] at (51.81796,0.651724) {Three seeds\\native};
\fill[palmblue] (16.77107,1.470864) circle[radius=2pt];
\node[above right,align=left] at (16.77107,1.470864) {Single\\scales and reflection};
\fill[palmgreen] (81.72337,1.752777) circle[radius=2.8pt];
\node[below left,align=right,text=palmgreen] at (81.72337,1.752777) {Three seeds\\scales and reflection};
\end{tikzpicture}
\caption{Measured accuracy and prediction cost of four EoMT inference configurations. The selected ensemble yields the highest validation score, while the transformed single model is much faster. The graph begins at 59.2\pct\ on the vertical axis to show the small differences clearly.}
\label{fig:cost}
\end{figure}

The selected ensemble improves mean IoU by 1.283 percentage points over the native single checkpoint and by 0.282 points over the transformed single checkpoint. Prediction time increases from 16.77 to 81.72 seconds for 84 images, a factor of 4.87. Figure~\ref{fig:cost} shows the tradeoff. Peak allocated inference memory remains about 2.70 GiB because models are processed sequentially. The highest scoring variant is selected because its runtime remains feasible for the challenge. The transformed single checkpoint offers a faster option at 60.671\pct\ mean IoU.

Combining EoMT and the linear DINOv3 system does not improve on the best native EoMT checkpoint for any tested weight. Reflection alone and the tested overlapping window approach also reduce mean IoU. These results caution against treating ensemble diversity or additional transformations as automatic improvements. The observed benefit belongs to the specific three seed, three scale, reflection configuration evaluated here.

\subsection{Class level behavior}
The selected prediction performs strongly on large scene regions. Sky reaches 98.88\pct\ IoU, Road 95.07\pct, Building 94.38\pct, Void 92.65\pct, Car 91.69\pct, and Water Surface 91.24\pct. Table~\ref{tab:classes} in the appendix reports every category. The average over the eight rare categories is 32.044\pct, compared with 31.103\pct\ for the best native single checkpoint.

Some gains are concentrated in difficult categories. Parking Lot increases from 28.41\pct\ to 39.73\pct\ and Railing from 0.38\pct\ to 11.23\pct. The ensemble also improves Sidewalk, Pedestrian, Truck, Infrastructure Box, and Parking Barrier. Other categories lose accuracy. Motorcycle decreases by 7.83 percentage points, Bus by 4.91, and Driver by 2.84. Bicycle remains weak at 15.86\pct\ and Stairs at 13.92\pct. These tradeoffs are visible even though the overall metric improves.

Overpass and Pruned Tree have zero IoU in the final prediction. Their extremely small validation support limits what can be inferred from this result, but they remain part of the official mean. Success on broad scene structure coexists with unresolved small and rare category errors.

\subsection{Qualitative results}
\begin{figure}[htbp]
\centering
\includegraphics[width=\linewidth]{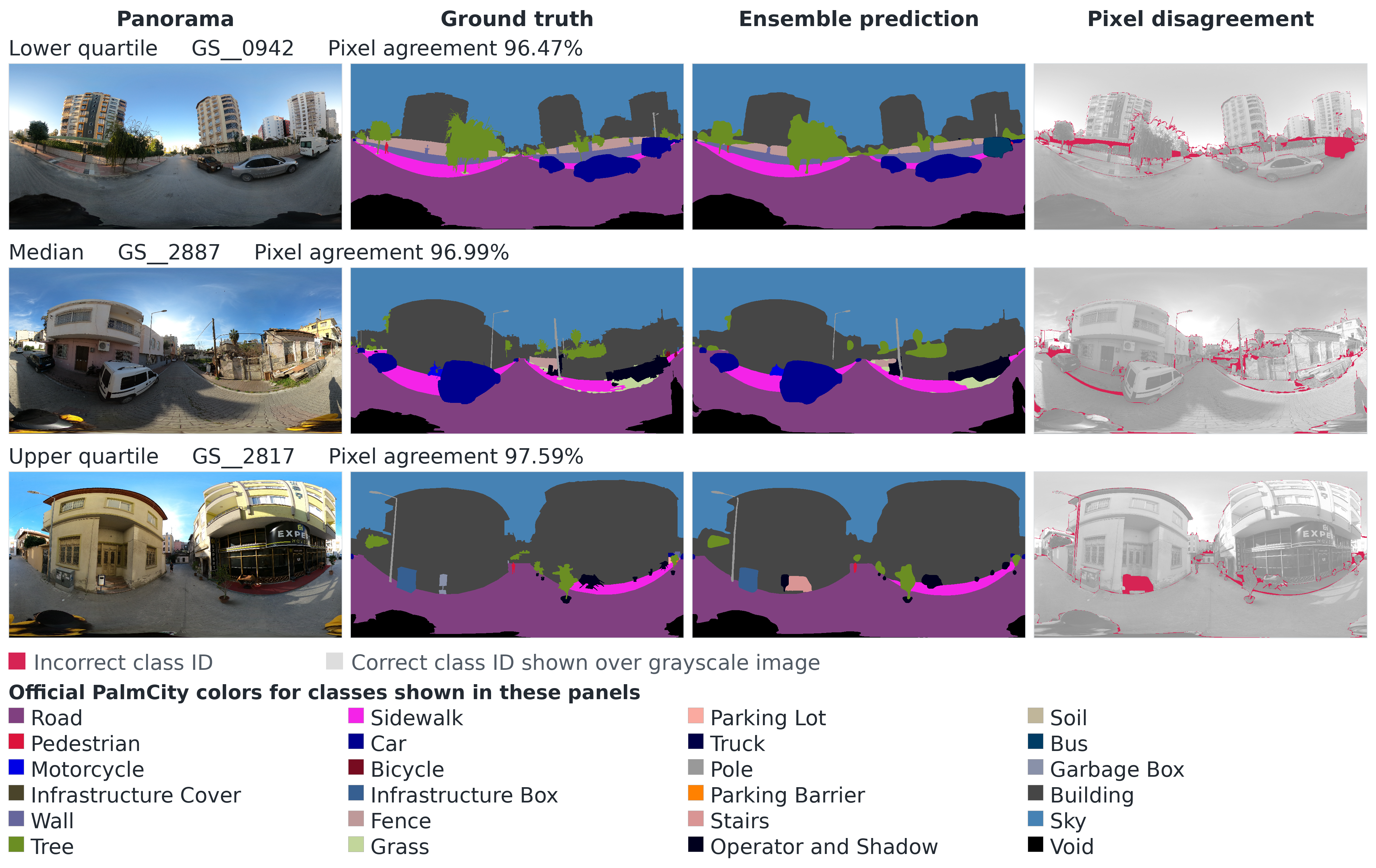}
\caption{Real validation examples nearest the lower quartile, median, and upper quartile of pixel agreement across all 84 images. This selection uses pixel agreement rather than class balanced IoU. Predictions are the saved outputs of the selected three seed ensemble with scales and reflection. Masks use the official PalmCity colors. Pink marks incorrect class identifiers, including Void.}
\label{fig:qualitative}
\end{figure}

Figure~\ref{fig:qualitative} compares real panoramas, reference annotations, and the saved predictions of the selected ensemble. The examples are selected by ranking all 84 validation images by the fraction of correctly labelled pixels and taking the nearest lower quartile, median, and upper quartile. This diagnostic gives an explicit selection rule for the illustrations. It differs from the official mean IoU because large regions contribute more heavily to pixel agreement. The examples retain much of the broad scene structure, while the disagreement maps reveal boundary errors and missed small objects.

\begin{figure}[htbp]
\centering
\includegraphics[width=\linewidth]{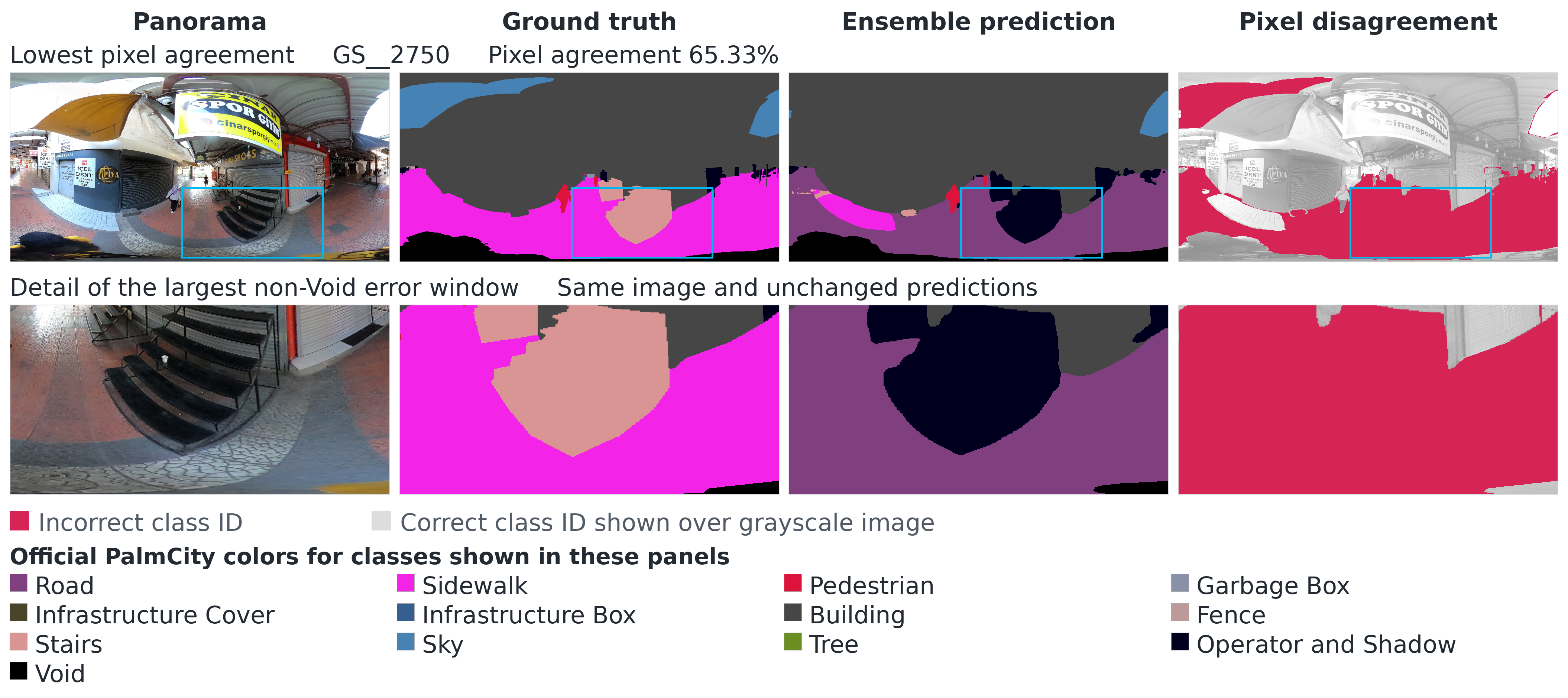}
\caption{The validation image with the lowest pixel agreement, \code{GS\_\_2750}. The blue box marks the detail shown below, selected as the $384\times192$ window with the most incorrect pixels outside reference Void regions. The shaded walkway is largely predicted as Road instead of Sidewalk, while the stairs are largely assigned to Operator and Shadow. Pink marks all incorrect class identifiers.}
\label{fig:failure}
\end{figure}

Figure~\ref{fig:failure} shows the validation image with the lowest pixel agreement, 65.33\pct. A shaded walkway is largely labelled as Road instead of Sidewalk, and the stairs are largely assigned to Operator and Shadow. The enlarged detail is the $384\times192$ window with the largest number of incorrect pixels whose reference category is not Void. The displayed reference masks and predictions are unchanged. This example makes the remaining ambiguity between street surfaces and small structures visible alongside the stronger results.

The illustrated images and reference annotations are credited to PalmCity \citep{iban2026palmcity} and are shown for academic illustration under its published noncommercial research and academic use conditions.

\subsection{Hidden test submission}
The final archive contains exactly 249 PNG files with the official test basenames. Every file is a single channel image at $1024\times512$ resolution, and every pixel contains an integer class identifier from 0 to 31. An independent audit checks these conditions and confirms that the archive contains the outputs of the three selected checkpoints. The archive is 1.82 MiB.

Test prediction takes 242.19 seconds for the 249 images. Total runtime is 251.49 seconds after including checkpoint loading and provenance checks. Peak allocated memory is 2.7015 GiB and peak reserved memory is 3.6133 GiB on one RTX 5090. This is a measured whole split result rather than a forward pass throughput estimate.

Table~\ref{tab:leaderboard} records the leaderboard snapshot associated with the submitted archive, independently retrieved from the public Codabench API \citep{codabench2026}. Submission 961441 by \code{yunusserhat}, dated 5 October 2026 at 09.15 Istanbul time, has 57.08\pct\ mean IoU and 67.96\pct\ mean F1. It leads the next listed entry by 11.05 percentage points in mean IoU and 10.98 points in mean F1 when calculated from the full precision API values. The ranking refers to this dated snapshot. It does not compare every possible method or future submission.

\begin{table}[htbp]
\centering
\caption{PalmCity Test Task leaderboard snapshot dated 5 October 2026. Values are percentages. The table identifies the submitted system without inferring the training procedure of other participants.}
\label{tab:leaderboard}
\begin{tabular}{rlrrr}
\toprule
Rank & Participant & Submission & mIoU & mF1\\
\midrule
1 & yunusserhat & 961441 & \textbf{57.08} & \textbf{67.96}\\
2 & onurcbayrak & 945126 & 46.02 & 56.99\\
3 & MaxwellEQ & 950834 & 33.54 & 40.93\\
\bottomrule
\end{tabular}
\end{table}

The hidden test mean IoU is 3.88 percentage points below the selected public validation score. The mean F1 difference is 3.20 points. The leaderboard score was obtained only after the final inference setting and archive were fixed. This snapshot did not remain final. Another participant's entry of 20 October 2026 reached 60.55\pct\ mean intersection over union. Section~\ref{sec:second} describes the second stage that followed.

\section{A second stage, 28 to 30 October 2026}
\label{sec:second}
\subsection{Starting point and selection rules}
The second stage starts from the recorded first stage. The three released models, downloaded from their fixed Hugging Face revision, reproduce the selected validation score of 60.95\pct\ exactly. All runs use the same dependency lock, validation split, and metric. Every EoMT variant keeps the first stage budget of 6351 updates with an effective batch of four and ten validation checks, and the checkpoint with the highest validation mean IoU is retained. The control E0 is the first stage EoMT run with seed 42, which reaches 57.970\pct.

\begin{figure}[htbp]
\centering
\begin{tikzpicture}[node distance=7mm and 11mm, every node/.style={font=\small}, stage/.style={draw=palmgreen,rounded corners=2pt,align=center,text width=6.7cm,minimum height=15mm,inner sep=5pt}, flow/.style={-{Stealth[length=2mm]},thick,palmgreen}]
\node[stage] (a) {First stage ensemble reproduced\\60.95\pct\ validation, same lock and split};
\node[stage,right=of a] (b) {Recipe and system candidates\\E1 to E3, P1, and C1 to C4};
\node[stage,below=of b] (c) {First chain with H1 seeds\\Three H1 models kept, 63.75\pct};
\node[stage,left=of c] (d) {Second chain with scales and H15\\Three H1 models and H15 kept, 64.53\pct};
\node[stage,below=of d,xshift=3.9cm,text width=14.6cm] (e) {Selected recipes retrained on 581 training and validation images\\Hidden test 60.25\pct\ on 29 October and 61.54\pct\ on 30 October 2026};
\draw[flow] (a)--(b);\draw[flow] (b)--(c);\draw[flow] (c)--(d);\draw[flow] (d.south)--(d.south |- e.north);
\end{tikzpicture}
\caption{The second stage. Each chain fixed its selection rule before its results existed, and both rules keep the highest validation mean IoU. Validation scores belong to models trained on the training split only. Hidden test labels were never used.}
\label{fig:stage2}
\end{figure}
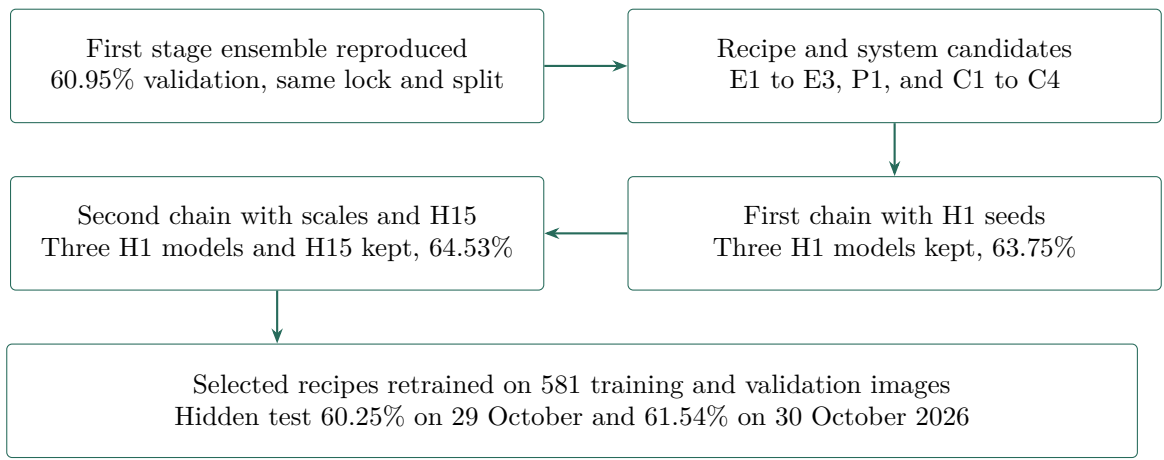

The runs were organized in two chains of scripts. Each chain wrote its selection rule into the script before the corresponding results existed and stored each decision in a separate file. Both rules choose the candidate with the highest validation mean IoU. The paired bootstrap of Section~\ref{sec:bootstrap} is reported for each choice but does not decide it. Figure~\ref{fig:stage2} summarizes the sequence. Hidden test labels were never available, and the second stage made two test submissions, each after a chain had fixed its archive.

\subsection{Recipe changes and further systems}
Table~\ref{tab:stage2runs} lists every second stage training run on the training split. E1 adds two augmentations to the reflection of the first stage. With probability 0.5 it takes a crop with the $2\mathbin{:}1$ aspect ratio covering 50\pct\ to 100\pct\ of the panorama area and resizes it to the input size. With probability 0.8 it changes brightness, contrast, saturation, and hue with strengths 0.2, 0.2, 0.2, and 0.05 in random order. Masks follow the crop with nearest neighbor resampling, and validation images are not augmented. E2 draws training images with weights that grow with the inverse frequency of the rare classes they contain, using power 0.5, a maximum weight of 5, and at least 16 pixels per class. Road, Building, Sky, and Void do not count toward rarity but remain in the loss and the metric. E3 combines E1 and E2. P1 raises the number of points used for query matching from 12544 to 50176. Its run stopped after 5000 of the planned 6351 updates and is reported as an incomplete run.

Four further systems change the starting point. C1 is SegFormer B5 initialized from its Cityscapes checkpoint \citep{xie2021segformer,cordts2016cityscapes}. C2 repeats Mask2Former with a longer schedule. C3 is the YOLO26x semantic segmentation model initialized from Cityscapes \citep{jocher2026yolo26}. C4 is EoMT initialized from a DINOv3 EoMT Cityscapes checkpoint released through LightlyTrain \citep{lightly2025train}. C1 and C2 receive 7500 updates and C3 receives 10000 updates, so these runs do not follow the equal computation protocol of the first stage.

\begin{table}[htbp]
\centering
\small
\setlength{\tabcolsep}{4.5pt}
\caption{Second stage runs on the training split. Scores are percentages from the checkpoint with the highest of ten validation checks. Rare IoU is the mean over the eight rare categories of Table~\ref{tab:pilots}. Minutes measure training and validation computation. GiB is peak allocated training memory. E0 is the first stage EoMT run with seed 42.}
\label{tab:stage2runs}
\begin{tabular}{llrrrrrrr}
\toprule
Run & Input & Seed & mIoU & mF1 & Rare IoU & Epoch & Minutes & GiB\\
\midrule
E0 & $512\times1024$ & 42 & 57.970 & 68.576 & 31.336 & 26 & 41.0 & 10.58\\
E1 & $512\times1024$ & 42 & 60.547 & 71.256 & 33.801 & 41 & 41.4 & 10.58\\
E1 & $512\times1024$ & 123 & 59.118 & 69.759 & 31.657 & 51 & 41.0 & 10.58\\
E1 & $512\times1024$ & 2026 & 59.099 & 69.536 & 32.398 & 51 & 43.9 & 10.58\\
E2 & $512\times1024$ & 42 & 58.667 & 68.925 & 30.207 & 51 & 41.1 & 10.58\\
E3 & $512\times1024$ & 42 & 58.911 & 69.677 & 30.052 & 36 & 41.1 & 10.58\\
P1, incomplete & $512\times1024$ & 42 & 58.638 & 68.961 & 29.729 & 36 & 32.8 & 10.68\\
C1 SegFormer B5 & $512\times1024$ & 42 & 50.846 & 61.665 & 18.317 & 54 & 35.3 & 6.02\\
C2 Mask2Former & $512\times1024$ & 42 & 54.586 & 65.185 & 24.155 & 42 & 51.8 & 9.37\\
C3 YOLO26x & $512\times1024$ & 42 & 48.384 & 57.914 & 9.815 & 41 & 14.9 & 3.40\\
C4 EoMT & $512\times1024$ & 42 & 57.644 & 68.203 & 24.878 & 36 & 47.1 & 10.71\\
H1 & $768\times1536$ & 42 & 61.496 & 71.831 & 30.419 & 46 & 94.8 & 19.28\\
H1 & $768\times1536$ & 123 & 62.288 & 72.079 & 34.494 & 21 & 95.0 & 19.29\\
H1 & $768\times1536$ & 2026 & 63.557 & 73.244 & 36.307 & 41 & 95.1 & 19.29\\
H1 & $768\times1536$ & 7 & 61.924 & 72.165 & 34.652 & 31 & 95.0 & 19.28\\
H1 & $768\times1536$ & 99 & 62.215 & 71.669 & 36.558 & 51 & 94.8 & 19.28\\
H15 & $896\times1792$ & 42 & 62.706 & 72.240 & 34.467 & 51 & 143.5 & 26.14\\
\bottomrule
\end{tabular}
\end{table}

E1 gives the largest single model gain at $512\times1024$ pixels. Its seed 42 run reaches 60.547\pct, compared with 57.970\pct\ for the same seed in E0, and its three seeds average 59.59\pct\ against 58.79\pct\ for the three first stage seeds. E2 reaches 58.667\pct\ but lowers the mean IoU of the eight rare categories from 31.336\pct\ to 30.207\pct\ for the same seed, so it does not reach its aim. E3 and the incomplete P1 lie between E0 and E1. None of the four further systems reaches E0, and C4 comes closest with 57.644\pct.

\subsection{Seeds and input size}
At $512\times1024$ pixels the patch grid has $32\times64$ cells, and small objects cover few of them. H1 trains the E1 recipe at $768\times1536$ pixels and H15 at $896\times1792$ pixels, with the same updates and schedule. The panorama is resampled to the input size, so these runs see the same image content on a finer grid of $48\times96$ or $56\times112$ cells. H1 reaches between 61.496\pct\ and 63.557\pct\ over five seeds, with a mean of 62.30\pct, against 59.59\pct\ for the three E1 seeds. The single H15 run reaches 62.706\pct. One seed cannot separate the effect of the larger input from seed variation, because the five H1 seeds already span 2.06 percentage points.

The larger inputs cost time and memory. An H1 run needs between 94.8 and 95.1 minutes of training and validation computation on one RTX 5090 and the H15 run 143.5 minutes, compared with 41.0 to 43.9 minutes for E1. Peak allocated training memory grows from 10.58 GiB to 19.29 GiB and 26.14 GiB. An input of $1024\times2048$ pixels did not fit in the 32 GB memory of one card. A short benchmark ran out of memory after 24.20 GiB of allocation, and the Transformers implementation of EoMT in the locked environment does not support gradient checkpointing.

\subsection{Ensembles and test time scales}
Table~\ref{tab:stage2ens} lists the validation ensembles, and Table~\ref{tab:stage2pairs} gives the paired comparisons that guided the two chains. With the first stage transforms, the three E1 seeds reach 62.04\pct, 1.08 points above the first stage ensemble, although the 95\pct\ interval of the difference runs from $-0.12$ to $+2.54$. Adding the H1 model with seed 42 raises the score to 62.73\pct. The first chain then trained the H1 seeds 123 and 2026 and compared the three H1 models alone with all six models. By the written rule the three H1 models were kept, with 63.75\pct\ against 63.39\pct. The bootstrap gives this choice a positive share of 0.944 and an interval from $-0.13$ to $+1.12$, so the data support it only weakly.

\begin{table}[htbp]
\centering
\small
\setlength{\tabcolsep}{4.5pt}
\caption{Validation ensembles of the second stage. Members are trained on the training split only and receive equal weights, and every scale is used with and without reflection. Scales refer to the input size of each model. Seconds measure prediction for the 84 images, and GiB is peak allocated inference memory. Row A is the first stage ensemble measured again in the second stage. Row K is the selected composition.}
\label{tab:stage2ens}
\begin{tabular}{lllrrrrr}
\toprule
Row & Members & Scales & mIoU & mF1 & Rare IoU & Seconds & GiB\\
\midrule
A & First stage ensemble & 0.75, 1, 1.25 & 60.953 & 71.165 & 32.044 & 93.12 & 2.70\\
B & E1, three seeds & 0.75, 1, 1.25 & 62.035 & 72.460 & 35.460 & 83.79 & 2.70\\
C & H1, seed 42 & 0.75, 1, 1.25 & 62.340 & 72.604 & 31.269 & 38.22 & 3.77\\
D & E1 three seeds, H1 seed 42 & 0.75, 1, 1.25 & 62.732 & 72.958 & 35.741 & 123.97 & 3.77\\
E & E1 and H1, three seeds each & 0.75, 1, 1.25 & 63.393 & 73.445 & 36.230 & 230.22 & 3.77\\
F & H1, three seeds & 0.75, 1, 1.25 & 63.754 & 73.641 & 35.799 & 148.67 & 3.77\\
G & H1, three seeds & 0.75, 1, 1.25, 1.5 & 63.831 & 73.497 & 35.770 & 184.12 & 4.62\\
H & H1, three seeds & 0.5, 0.75, 1, 1.25 & 63.574 & 73.657 & 34.858 & 130.27 & 3.77\\
I & H1, three seeds & 1, 1.25, 1.5 & 64.014 & 73.476 & 36.266 & 170.74 & 4.62\\
J & H1, five seeds & 1, 1.25, 1.5 & 64.039 & 73.320 & 37.238 & 299.43 & 4.62\\
K & H1 three seeds, H15 & 1, 1.25, 1.5 & \textbf{64.533} & 73.672 & 36.963 & 264.93 & 5.63\\
L & H1 five seeds, H15 & 1, 1.25, 1.5 & 64.384 & 73.566 & 37.428 & 363.70 & 5.63\\
\bottomrule
\end{tabular}
\end{table}

\begin{table}[htbp]
\centering
\small
\caption{Paired image bootstrap comparisons between the rows of Table~\ref{tab:stage2ens}. The difference is the first row minus the second in percentage points of mean IoU, with the 2.5 and 97.5 percentiles of 5000 replicates. Share is the fraction of replicates with a positive difference.}
\label{tab:stage2pairs}
\begin{tabular}{lrcr}
\toprule
Comparison & Difference & 95\pct\ interval & Share\\
\midrule
B and A & $+1.08$ & $-0.12$ to $+2.54$ & 0.958\\
C and B & $+0.30$ & $-0.52$ to $+2.16$ & 0.836\\
D and B & $+0.70$ & $+0.41$ to $+1.17$ & 0.999\\
E and D & $+0.66$ & $+0.30$ to $+1.11$ & 1.000\\
F and D & $+1.02$ & $+0.25$ to $+2.15$ & 0.995\\
F and E & $+0.36$ & $-0.13$ to $+1.12$ & 0.944\\
G and F & $+0.08$ & $-0.25$ to $+0.48$ & 0.739\\
H and F & $-0.18$ & $-0.69$ to $+0.35$ & 0.235\\
I and F & $+0.26$ & $-0.50$ to $+0.99$ & 0.768\\
J and I & $+0.02$ & $-0.64$ to $+0.76$ & 0.549\\
K and I & $+0.52$ & $+0.02$ to $+1.02$ & 0.979\\
L and I & $+0.37$ & $-0.20$ to $+0.98$ & 0.880\\
K and A & $+3.58$ & $+2.31$ to $+5.59$ & 1.000\\
\bottomrule
\end{tabular}
\end{table}

The second chain first compared four scale sets for the three H1 models. Scales 1, 1.25, and 1.5 gave the highest score, 64.01\pct, a difference of $+0.26$ with an interval from $-0.50$ to $+0.99$. It then trained the H1 seeds 7 and 99 and the H15 model. Five H1 models give 64.04\pct, practically the same as three. Adding the H15 model to the three H1 models gives 64.53\pct, the highest score, with a difference of $+0.52$, an interval from $+0.02$ to $+1.02$, and a positive share of 0.979. Five H1 models with the H15 model give 64.38\pct. Against the first stage ensemble, the selected composition gains 3.58 points with an interval from $+2.31$ to $+5.59$ and is higher in all 5000 replicates. The gain has a cost. The selected composition needs 264.93 seconds and 5.63 GiB for the 84 validation images, compared with 93.12 seconds and 2.70 GiB for the first stage ensemble measured again in the second stage.

\subsection{Negative results}
Several alternatives lowered the validation score, as Table~\ref{tab:negative} shows. Semantic Segment Anything assigns classes to the masks of Segment Anything by voting over an existing semantic prediction \citep{chen2023ssa,kirillov2023sam}. It was applied to the first stage ensemble with the ViT H model and the mask generator settings of that project, in three variants. All three lowered mean IoU, by 0.09 to 0.92 points. A YOLOv8x instance segmentation model \citep{jocher2023ultralytics}, initialized from COCO \citep{lin2014coco} and fine tuned for 100 epochs at 1024 pixels on 11 object classes derived from the training masks, was used to overwrite object pixels of the E1 ensemble. The three overlays lowered the score by 0.65 to 3.12 points. With all classes at confidence 0.5, Driver lost 47.4 points, Truck 17.0, and Motorcycle 16.0. Adding C2 or C3 to the first stage ensemble with weights of 0.1, 0.25, or 0.5 fixed in advance reduced the score in every case, as did an equal ensemble of E2 and C2. These results agree with the first stage observation that combining EoMT with weaker systems does not help.

\begin{table}[htbp]
\centering
\small
\setlength{\tabcolsep}{4.5pt}
\caption{Alternatives that reduced the validation score. Base names the baseline row of Table~\ref{tab:stage2ens}. Differences, intervals, and positive shares follow Table~\ref{tab:stage2pairs}. The C3 combinations have no bootstrap record, so only their score difference is given.}
\label{tab:negative}
\begin{tabular}{llrrrcr}
\toprule
Variant & Base & mIoU & mF1 & Difference & 95\pct\ interval & Share\\
\midrule
SSA voting & A & 60.554 & 70.751 & $-0.40$ & $-1.23$ to $+0.16$ & 0.076\\
SSA, weak classes kept & A & 60.865 & 71.239 & $-0.09$ & $-0.83$ to $+0.39$ & 0.289\\
SSA, small masks skipped & A & 60.028 & 70.373 & $-0.92$ & $-2.43$ to $+0.06$ & 0.032\\
YOLOv8, all classes, 0.5 & B & 58.917 & 69.490 & $-3.12$ & $-5.53$ to $-0.33$ & 0.010\\
YOLOv8, all classes, 0.7 & B & 59.440 & 70.204 & $-2.60$ & $-4.51$ to $-0.03$ & 0.022\\
YOLOv8, four classes, 0.5 & B & 61.387 & 71.943 & $-0.65$ & $-1.65$ to $-0.08$ & 0.013\\
A and C2, weight 0.10 & A & 60.918 & 71.091 & $-0.03$ & $-0.22$ to $+0.38$ & 0.444\\
A and C2, weight 0.25 & A & 60.199 & 70.212 & $-0.75$ & $-1.36$ to $+0.17$ & 0.070\\
A and C2, weight 0.50 & A & 58.664 & 68.306 & $-2.29$ & $-3.42$ to $-0.59$ & 0.002\\
A and C3, weight 0.10 & A & 60.832 & 70.950 & $-0.12$ &  & \\
A and C3, weight 0.25 & A & 59.968 & 69.856 & $-0.99$ &  & \\
A and C3, weight 0.50 & A & 56.220 & 65.821 & $-4.73$ &  & \\
E2 and C2, equal weights & A & 58.895 & 68.671 & $-2.06$ & $-3.24$ to $-0.15$ & 0.015\\
\bottomrule
\end{tabular}
\end{table}

\subsection{Final training on training and validation images}
\label{sec:trainval}
The selected recipes were retrained on a manifest whose training list holds the 497 training and the 84 validation images. The 6351 updates then span 44 epochs instead of 51. No intermediate validation is scheduled, and the weights after the last update are used. This choice was fixed before the H1 seeds 123 and 2026 had finished, when the only H1 run ended 0.07 points below its best check. Over all five H1 runs on the training split, the last check lies between 0 and 1.45 points below the best one, so the last weights may cost some accuracy. With the validation images inside training, no better selection was available. The single validation check at the end gives between 80.796\pct\ and 82.163\pct\ mean IoU, which only shows that training completed.

Test prediction with the four retrained models takes 1070.92 seconds for the 249 images, or 1088.71 seconds including checkpoint loading and provenance checks, on one RTX 5090. Peak allocated memory is 5.63 GiB, compared with 242.19 seconds, 251.49 seconds, and 2.70 GiB for the first stage system. Both second stage archives contain exactly 249 single channel PNG files, and the final archive is 1.91 MiB.

\subsection{Hidden test results}
The second stage made two test submissions. The first, on 29 October 2026, used the three retrained H1 models with scales 0.75, 1, and 1.25 and received 60.25\pct\ mean intersection over union and 70.82\pct\ mean F1, below the entry of 60.55\pct. The second, on 30 October 2026, used the selected four models and received 61.54\pct\ and 71.93\pct. Table~\ref{tab:leaderboard2} records the public snapshot retrieved on 30 October 2026 \citep{codabench2026second}, which lists the best entry of each participant. The submission leads the next entry by 0.98 points in mean IoU and 0.51 points in mean F1, calculated from the full precision values.

\begin{table}[htbp]
\centering
\caption{PalmCity Test Task leaderboard snapshot dated 30 October 2026. Values are percentages. The table identifies the submitted system without inferring the training procedure of other participants.}
\label{tab:leaderboard2}
\begin{tabular}{rlrrr}
\toprule
Rank & Participant & Submission & mIoU & mF1\\
\midrule
1 & yunusserhat & 1020616 & \textbf{61.54} & \textbf{71.93}\\
2 & korayaykor & 1005153 & 60.55 & 71.42\\
3 & formid & 1018170 & 47.21 & 58.98\\
4 & onurcbayrak & 945126 & 46.02 & 56.99\\
5 & MaxwellEQ & 950834 & 33.54 & 40.93\\
\bottomrule
\end{tabular}
\end{table}

The validation score of the training only counterpart exceeds the test score by 2.99 points, compared with 3.88 points in the first stage. The two second stage numbers belong to different model instances, so this difference is not a measured generalization gap of the submitted models.

\subsection{Class level changes}
Table~\ref{tab:classes} in the appendix adds the second stage ensembles. Relative to the first stage ensemble, the selected composition improves Garbage Box by 15.85 points, Traffic Sign by 14.61, Parking Barrier by 12.92, Pedestrian by 10.31, Traffic Light by 9.30, Bicycle by 8.98, and Pole by 8.36. The mean over the eight rare categories rises from 32.04\pct\ to 36.96\pct. Losses remain. Railing falls from 11.23\pct\ to 2.41\pct, Truck by 6.99 points, and Parking Lot by 3.92 points. Stairs stays at 13.29\pct, and Overpass and Pruned Tree remain at zero. Traffic Light appears in only one validation image, so its change is weak evidence.

\begin{figure}[htbp]
\centering
\includegraphics[width=\linewidth]{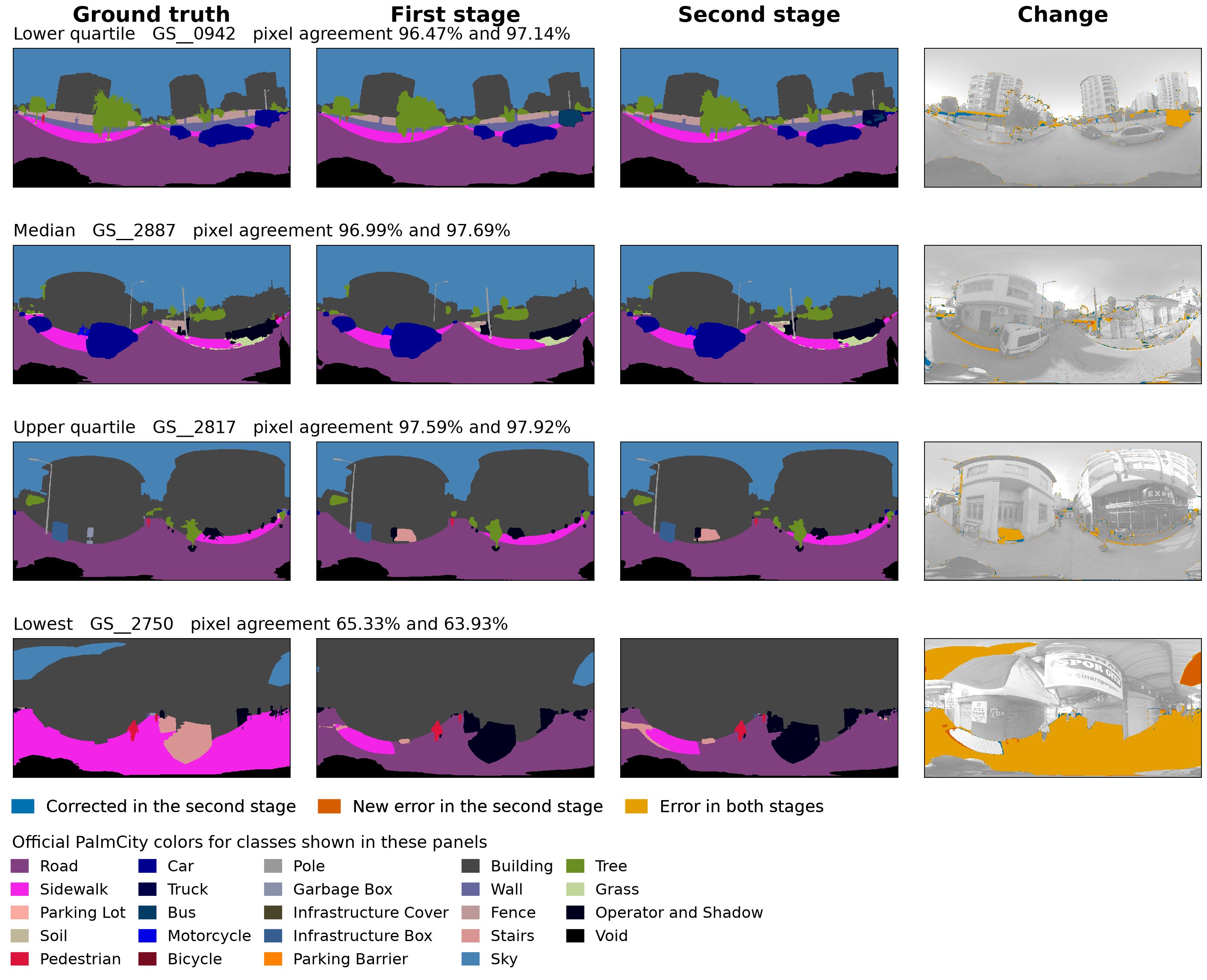}
\caption{The validation images of Figures~\ref{fig:qualitative} and~\ref{fig:failure} with the saved predictions of the first stage ensemble and of the selected second stage composition trained on the training split only. The change panel marks pixels corrected by the second stage, new errors, and errors of both stages over the grayscale panorama. Pixel agreement is given for the first and the second stage. Masks use the official PalmCity colors.}
\label{fig:stage2qual}
\end{figure}

Figure~\ref{fig:stage2qual} repeats the validation images of Figures~\ref{fig:qualitative} and~\ref{fig:failure} with the first stage ensemble and the training only counterpart of the selected composition. Pixel agreement rises from 96.47\pct\ to 97.14\pct, from 96.99\pct\ to 97.69\pct, and from 97.59\pct\ to 97.92\pct\ for the three quartile examples. On the image with the lowest agreement it falls from 65.33\pct\ to 63.93\pct. The shaded walkway is still labelled as Road, and the stairs are still assigned to Operator and Shadow. As before, the images and annotations are credited to PalmCity \citep{iban2026palmcity}.

\section{Discussion}
\subsection{What the comparison supports}
The evidence favors EoMT with the recorded DINOv3 ViT L checkpoint and PalmCity recipe among the systems tested. It wins the eligible pilot and remains ahead of the other confirmed candidate across three independent seeds. The final probability ensemble with scale and reflection transforms increases validation accuracy further. These results make it a defensible choice for this challenge and computation budget.

The experiment also supports a more general procedural lesson. Correct class semantics, suitable source initialization, budget calibration, and output verification matter alongside the architecture name. Query probability normalization and rectangular mask alignment affect the predicted pixels, while CPU and GPU model transfers affect the measured runtime. These implementation choices accompany the reported score so that another researcher can recover the same procedure.

The systems have different pretrained data, decoders, and loss functions. EoMT begins from an ADE20K segmentation checkpoint, while the linear DINOv3 candidate begins from a representation checkpoint with a new decoder. Larger or more suitable decoders for DINOv3 may narrow the difference. The pilot budget gives a practical answer about transfer configurations under limited search. Isolating architectural contributions would require matched pretraining and additional controlled comparisons.

Only two candidates received longer training in the first stage. The second stage gave Mask2Former a longer schedule, which reached 54.586\pct, still below every EoMT run. Broader augmentation helped, and rare class image sampling did not raise the rare categories. Longer EoMT schedules and additional decoder comparisons remain open questions.

The second stage adds evidence about where the gains came from under a fixed update budget. Within EoMT, color and crop augmentation and larger inputs gave the clearest improvements. Averaging three seeds per recipe helped, while two further H1 seeds added almost nothing. Systems with other architectures or initializations did not add useful diversity in any tested weighting, and post processing with Segment Anything masks or instance masks reduced accuracy. These observations concern this dataset, this budget, and these recipes.

\subsection{Implications for urban image analysis}
The high scores for road, sky, building, tree, and water suggest that the system can produce useful broad scene descriptions on imagery similar to the benchmark. However, a high mean IoU does not certify the reliability of a derived environmental indicator. For example, a measure concerned with bicycle infrastructure, access barriers, or small traffic controls may be sensitive to the classes that remain weak. Error inspection must be tied to the intended measurement rather than to a single overall ranking. The second stage improved several small object classes that matter for inventories of street furniture, such as poles, signs, benches, and garbage boxes, while railings and stairs remained unreliable.

Visible greenness and environmental quality have been studied through different sensors and sampling strategies \citep{helbich2021greenness,biljecki2021review}. PalmCity segmentation could be one input to a subsequent urban analysis. Linking it to health, urban redesign, or policy would require a separate study of acquisition conditions, geographical sampling, temporal coverage, and task specific error. A benchmark trained model should be evaluated again before use in a new city or comparison between groups of residents.

Decoder design and adaptation also deserve attention across image domains. The author's earlier ATTransUNet work combines attention and transformer components for building segmentation from aerial imagery and laser data \citep{bicakci2023attransunet}. Panoramic street parsing presents a complementary problem with different image geometry, annotation needs, and transfer conditions.

Temporal consistency provides another potential extension when connected panoramas or street video become available. Recent VidEoMT work extends encoder focused mask prediction to video \citep{norouzi2026videomt}. This direction could be evaluated on future data with temporal connections, while the present system processes each panorama independently.

\subsection{Limits of the evidence}
The 84 validation images select checkpoints, candidates for confirmation, individual seeds, and the final inference variant. This reuse creates selection dependence. The 0.282 percentage point advantage of the selected ensemble over the faster transformed single model may change on an independent sample. Three seed population deviations describe training variation, not confidence intervals over new streets. Visually similar views across the official split further limit claims of geographical independence. Route, GPS, and acquisition time metadata would be needed for a spatial evaluation. The validation and hidden test difference may reflect split composition, sampling variation, or repeated validation selection, but its causes cannot be isolated from aggregate scores.

The second stage strengthens these concerns. The same 84 images selected checkpoints, test time scales, and the final composition through two chains of comparisons, so its validation scores are optimistic as well. Several bootstrap intervals include zero, and the interval for adding the H15 model starts at $+0.02$ points. The submitted models were trained on the validation images and have no independent validation score. The hidden test score is the only independent measurement of the final system, and it was observed twice during the second stage. The second chain was approved right after the first second stage submission, with the aim of passing the leading entry. Its choices used only validation data and rules fixed in advance, and no setting was tuned on the hidden test score, but the decision to continue the search was informed by the leaderboard. The single H15 run cannot separate input size from seed variation. Finally, the second stage used 23.94 GPU hours of measured training and validation computation for 20 runs, compared with 6.34 hours for the 19 first stage runs, so the two stages are not compared at equal cost.

The results establish the highest score among the public entries in two dated leaderboard snapshots, with three entries on 5 October 2026 and five on 30 October 2026. The competition has no published end date. Architecture wide superiority, final competition victory, and generalization to other cities would require additional evidence, and no inference is made about the methods of other participants. The 19 first stage runs and the 20 second stage runs constitute a bounded search with adapted source recipes, rather than exhaustive optimization of each candidate.

\subsection{Public release}

The public code release lets another researcher prepare data, obtain permitted pretrained sources, train the selected systems, and generate a challenge archive in a separate workspace. It includes the dependency lock, configurations, result summaries, source identities, and commands for the experiment stages. The project code is released under the GNU General Public License version 3 only, identified as \code{GPL-3.0-only}. PalmCity images, dependencies, and third party pretrained weights retain their original conditions. Some evaluated sources have research or noncommercial restrictions, and DINOv3 has its own access conditions. Release 0.2.0 adds the second stage recipes, the augmentation and sampling options, the paired bootstrap, the manifest of training and validation images, and reproduction stages for the selected system. Applied to the original workspace, it reproduces the recorded bootstrap of the selected composition, the training and validation manifest, and the 16 second stage EoMT run configurations exactly. The code of C3, C4, and the two post processing methods is not part of the release, because it needs separate environments or packages under the AGPL license, and none of these methods was selected.

The published recipe allows repetition of the training procedure, while CUDA nondeterminism and hardware behavior can change a rerun. The training and inference commands document how the submitted predictions were produced.

\section{Conclusion}
This study documents two stages of a PalmCity challenge submission. In the first stage, approximately equal computation pilots, three seed confirmation of the leading candidates, and a complete inference comparison lead to a three seed EoMT DINOv3 ViT L ensemble with 60.95\pct\ validation and 57.08\pct\ hidden test mean IoU, first in the leaderboard snapshot of 5 October 2026. In the second stage, color and crop augmentation and larger inputs improve the single models, and a four model ensemble trained at 768 by 1536 and 896 by 1792 pixels and retrained on the training and validation images receives 61.54\pct\ hidden test mean IoU, first in the snapshot of 30 October 2026. The record also shows which alternatives failed, what the gains cost in training and inference time, and how much the validation split was reused. It provides a reproducible challenge solution and a basis for further street image segmentation research.

\section*{Data and code availability}
PalmCity is available through the dataset authors' \href{https://github.com/PalmCityDataset/palmcity}{official repository}. The training and inference code is available in the \href{https://github.com/yunusserhat/palmcity_challenge}{PalmCity challenge repository}, release 0.2.0. The three first stage models and the four second stage models are provided as native Transformers safetensors checkpoints in the \href{https://huggingface.co/yunusserhat/palmcity-eomt-dinov3-large}{first} and \href{https://huggingface.co/yunusserhat/palmcity-eomt-dinov3-large-v2}{second} Hugging Face model releases, with instructions for direct inference without retraining. The second stage models were trained on the training and validation images. The adapted weights retain their applicable source licences, including the DINOv3 terms supplied with each release. The manuscript and its source are maintained separately from the code repository.

\section*{Acknowledgements}
The author thanks the Geospatial Data Science Group at the University of Glasgow. The author also thanks the PalmCity dataset creators and challenge organizers for providing the benchmark and evaluation platform, Meta AI for DINOv3, and the developers of EoMT and the other models and software used in the experiments.

\clearpage
\appendix
\section{Complete class results}
\begingroup
\small
\begin{longtable}{rlrrrrrr}
\caption{Validation IoU for every official category. Single refers to EoMT seed 2026 at the native scale. Selected is the first stage three seed ensemble with scales and reflection, and the first difference compares it with Single. H1 is the second stage ensemble of three H1 seeds with scales 0.75, 1, and 1.25. Second is the selected second stage composition of Table~\ref{tab:stage2ens}, trained on the training split only, and the last difference compares it with Selected. Differences are percentage points calculated before rounding.}\label{tab:classes}\\
\toprule
ID & Class & Single & Selected & Difference & H1 & Second & Difference\\
\midrule
\endfirsthead
\toprule
ID & Class & Single & Selected & Difference & H1 & Second & Difference\\
\midrule
\endhead
\bottomrule
\endfoot
0 & Road & 94.81 & 95.07 & $+0.26$ & 95.11 & 95.21 & $+0.14$\\
1 & Sidewalk & 77.36 & 78.45 & $+1.10$ & 78.27 & 78.33 & $-0.12$\\
2 & Parking Lot & 28.41 & 39.73 & $+11.31$ & 38.96 & 35.81 & $-3.92$\\
3 & Soil & 76.21 & 78.94 & $+2.73$ & 81.45 & 81.60 & $+2.66$\\
4 & Pedestrian & 65.88 & 69.22 & $+3.34$ & 75.80 & 79.53 & $+10.31$\\
5 & Driver & 73.91 & 71.08 & $-2.84$ & 76.30 & 78.87 & $+7.79$\\
6 & Car & 91.26 & 91.69 & $+0.44$ & 90.36 & 92.10 & $+0.40$\\
7 & Truck & 69.58 & 72.47 & $+2.89$ & 62.11 & 65.48 & $-6.99$\\
8 & Bus & 80.41 & 75.51 & $-4.91$ & 79.69 & 81.24 & $+5.74$\\
9 & Motorcycle & 71.53 & 63.70 & $-7.83$ & 67.46 & 68.73 & $+5.03$\\
10 & Bicycle & 17.63 & 15.86 & $-1.77$ & 24.87 & 24.83 & $+8.98$\\
11 & Traffic Light & 44.34 & 44.85 & $+0.51$ & 51.67 & 54.15 & $+9.30$\\
12 & Traffic Sign & 43.49 & 43.09 & $-0.40$ & 58.14 & 57.70 & $+14.61$\\
13 & Pole & 59.52 & 61.56 & $+2.04$ & 67.27 & 69.92 & $+8.36$\\
14 & Garbage Box & 53.78 & 57.48 & $+3.70$ & 70.91 & 73.34 & $+15.85$\\
15 & Sitting Bench & 69.07 & 70.25 & $+1.18$ & 67.75 & 77.74 & $+7.49$\\
16 & Infrastructure Cover & 74.19 & 75.63 & $+1.44$ & 76.56 & 77.23 & $+1.60$\\
17 & Infrastructure Box & 62.14 & 65.28 & $+3.14$ & 67.57 & 69.75 & $+4.47$\\
18 & Parking Barrier & 51.90 & 54.83 & $+2.93$ & 63.15 & 67.75 & $+12.92$\\
19 & Building & 93.86 & 94.38 & $+0.52$ & 94.99 & 95.37 & $+0.99$\\
20 & Wall & 68.36 & 70.85 & $+2.50$ & 73.99 & 75.15 & $+4.29$\\
21 & Fence & 63.05 & 63.57 & $+0.52$ & 65.38 & 66.75 & $+3.18$\\
22 & Stairs & 12.54 & 13.92 & $+1.37$ & 25.37 & 13.29 & $-0.63$\\
23 & Railing & 0.38 & 11.23 & $+10.84$ & 7.67 & 2.41 & $-8.82$\\
24 & Overpass & 0.00 & 0.00 & $0.00$ & 0.00 & 0.00 & $0.00$\\
25 & Water Surface & 88.91 & 91.24 & $+2.33$ & 92.02 & 93.00 & $+1.76$\\
26 & Sky & 98.76 & 98.88 & $+0.12$ & 98.98 & 99.09 & $+0.21$\\
27 & Tree & 81.11 & 81.73 & $+0.62$ & 83.85 & 84.99 & $+3.26$\\
28 & Grass & 66.97 & 67.13 & $+0.16$ & 67.25 & 67.95 & $+0.81$\\
29 & Pruned Tree & 0.00 & 0.00 & $0.00$ & 0.00 & 0.00 & $0.00$\\
30 & Operator and Shadow & 38.09 & 40.23 & $+2.15$ & 44.99 & 45.45 & $+5.21$\\
31 & Void & 92.03 & 92.65 & $+0.63$ & 92.25 & 92.31 & $-0.35$\\
\end{longtable}
\endgroup

\section{Independent checks}
Independent training checks cover 19 runs, 190 validation evaluations, and 1162 training and validation input files. Independent inference checks directly decode 924 PNG files from the eleven validation variants and reconstruct their confusion matrices. The archive check covers all 249 members, PNG modes, native dimensions, class ranges, archive integrity, source images, and selected checkpoints. These checks confirm that the reported scores and the submitted masks correspond to the recorded experiments.

For the second stage, the records cover 20 training runs with their configurations and validation histories, 22 validation prediction sets with their reports and timing records, 24 paired bootstrap comparisons, the three decision files of the selection chains, and the strict checks of both submission archives. The bootstrap tool verifies each baseline prediction set against its stored confusion matrix before resampling.

\section{Reproduction outline}
\subsection{Direct inference with the released models}
The public repository separates project code from the data and output workspace. After cloning it, a reader prepares a project environment using the supplied setup script and its dependency lock. The released second stage models can then be applied to a local panorama or a directory of panoramas without downloading PalmCity or running training. Input panoramas must be twice as wide as they are high. The following command downloads the four trained checkpoints and applies the selected ensemble with scales and reflection.
\begingroup\small\begin{verbatim}
export PALMCITY_WORKSPACE=/absolute/user-owned/workspace
export PALMCITY_PYTHON=python3.12
bash scripts/setup.sh
bash scripts/verify.sh
HF_HUB_OFFLINE=0 bash scripts/run.sh python -m palmcity.hub \
  --download --input /absolute/path/panoramas \
  --output-dir outputs/challenge --mode challenge --colorize
\end{verbatim}\endgroup
The command writes class identifier PNGs and optional color visualizations into the selected workspace. The model release documents the source licences and the project helper preserves the rectangular mask geometry, Void category, and probability normalization used in the reported experiments. The faster single model options are also documented. Once the required weights are cached, inference can run offline. The first stage models remain available by adding \code{--model yunusserhat/palmcity-eomt-dinov3-large} and the fixed revision \code{addc5a5ef93a26c1d64c1fb23b8c1880ac11c22b} to the command.

\subsection{Repeating the first stage training}
A reader downloads PalmCity from the official source, creates and audits the split manifest, and obtains the pretrained checkpoint at its recorded repository revision. Access gated sources require the reader's own authorized account. The supplied selected configuration trains EoMT independently for seeds 42, 123, and 2026 with the recorded update and validation limits. After the environment setup above, the main training commands are
\begingroup\small\begin{verbatim}
bash scripts/reproduce.sh data
HF_HUB_OFFLINE=0 bash scripts/reproduce.sh prepare --download
bash scripts/reproduce.sh train
bash scripts/reproduce.sh val
bash scripts/reproduce.sh test
\end{verbatim}\endgroup

After training, the inference command uses the three best checkpoints, equal weights, scales 0.75, 1, and 1.25, and horizontal reflection. The submission command packages the test masks and rejects an incomplete archive. The essential inference settings are
\begingroup\small\begin{verbatim}
--weights 1 1 1
--scales 0.75 1 1.25
--hflip-tta
--split test
\end{verbatim}\endgroup
The repository documents the complete commands, expected split counts, checkpoint access conditions, memory checks, and output locations. Full comparison commands are also supplied for calibration, eligible pilots, confirmation, inference evaluation, and archive construction. A reader may reproduce the selected system alone or repeat the broader study. The original experiment records remain available alongside the portable training workflow.

\subsection{Repeating the second stage}
After the environment and data steps above, the second stage system is prepared, trained, evaluated, retrained, and packaged by
\begingroup\small\begin{verbatim}
HF_HUB_OFFLINE=0 bash scripts/reproduce.sh v2-prepare --download
bash scripts/reproduce.sh v2-train
bash scripts/reproduce.sh v2-val
bash scripts/reproduce.sh v2-trainval
bash scripts/reproduce.sh v2-test
\end{verbatim}\endgroup
The essential inference settings are
\begingroup\small\begin{verbatim}
--weights 1 1 1 1
--scales 1 1.25 1.5
--hflip-tta
--max-scaled-pixels 4000000
--split test
\end{verbatim}\endgroup
The repository also contains the recipes of all 16 second stage EoMT runs and the commands for the manifest of training and validation images and for the paired bootstrap.

\begingroup
\small
\setlength{\bibsep}{2pt}
\bibliographystyle{unsrtnat}
\providecommand{\doi}[1]{\href{https://doi.org/#1}{\nolinkurl{#1}}}
\renewcommand{\url}[1]{\href{#1}{Full text}}
\bibliography{references}
\endgroup
\end{document}